\documentclass[11pt,a4paper]{article}
\usepackage[hyperref]{acl2020}
\usepackage{booktabs}
\usepackage{times}
\usepackage{latexsym}
\usepackage{todonotes}

\usepackage[normalem]{ulem}
\usepackage{soul}

\usepackage{microtype}
\usepackage{enumitem}

\aclfinalcopy 
\def\aclpaperid{2806} 

\usepackage{subcaption}
\DeclareCaptionSubType * [alph]{table}
\newcommand{\verified}{verified} 
\newcommand{\verclaim}{VerClaim} 
\newcommand{\inputclaim}{input} 
\newcommand{\Inputclaim}{Input} 

\title{That is a Known Lie: Detecting Previously Fact-Checked Claims}

\author{Shaden Shaar$^1$, Nikolay Babulkov$^2$, Giovanni Da San Martino$^1$, Preslav Nakov$^1$\\
  $^1$Qatar Computing Research Institute, HBKU, Doha, Qatar \\
  $^2$Sofia University, Sofia, Bulgaria \\
  \texttt{\{sshar, gmartino, pnakov\}@hbku.edu.qa}\\\texttt{nbabulkov@gmail.com} \\}

\date{}

\begin{document}
\maketitle
\begin{abstract}
The recent proliferation of ``fake news'' has triggered a number of responses, most notably the emergence of several manual fact-checking initiatives. As a result and over time, a large number of fact-checked claims have been accumulated, which increases the likelihood that a new claim in social media or a new statement by a politician might have already been fact-checked by some trusted fact-checking organization, as viral claims often come back after a while in social media, and politicians like to repeat their favorite statements, true or false, over and over again. As manual fact-checking is very time-consuming (and fully automatic fact-checking has credibility issues), it is important to try to save this effort and to avoid wasting time on claims that have already been fact-checked. 
Interestingly, despite the importance of the task, it has been largely ignored by the research community so far. Here, we aim to bridge this gap. In particular, we formulate the task and we discuss how it relates to, but also differs from, previous work. We further create a specialized dataset, which we release to the research community. Finally, we present learning-to-rank experiments that demonstrate sizable improvements over state-of-the-art retrieval and textual similarity approaches.
\end{abstract}

\section{Introduction\label{sec:introduction}}

The year 2016 was marked by massive disinformation campaigns related to Brexit and the US Presidential Elections. While false statements are not a new phenomenon, e.g.,~yellow press and tabloids have been around for decades, this time things were notably different in terms of scale and effectiveness thanks to social media platforms, which provided both a medium to reach millions of users and an easy way to micro-target specific narrow groups of voters based on precise geographical, demographic, psychological, and/or political profiling. 

\noindent Governments, international organizations, tech companies, media, journalists, and regular users launched a number of initiatives to limit the impact of the newly emerging large-scale weaponization of \emph{disinformation}\footnote{In the public discourse, the problem is generally known as ``\emph{fake news}'', a term that was declared Word of the Year 2017 by Collins dictionary. Despite its popularity, it remains a confusing term, with no generally agreed upon definition. It is also misleading as it puts emphasis on (a)~the claim being false, while generally ignoring (b)~its intention to do harm. In contrast, the term \emph{disinformation} covers both aspects (a) and (b), and it is generally preferred at the EU level.} 
online. Notably, this included manual fact-checking initiatives, which aimed at debunking various false claims, with the hope to limit its impact, but also to educate the public that not all claims online are true. 

Over time, the number of such initiatives grew substantially, e.g.,~at the time of writing, the Duke Reporters' Lab lists 237 active fact-checking organizations plus another 92 inactive.\footnote{\url{http://reporterslab.org/fact-checking/}}
While some organizations debunked just a couple of hundred claims, others such as Politifact,\footnote{\url{http://www.politifact.com/}}
FactCheck.org,\footnote{\url{http://www.factcheck.org/}}
Snopes,\footnote{\url{http://www.snopes.com/}} 
and Full Fact\footnote{\url{http://fullfact.org/}}
have fact-checked thousands or even tens of thousands of claims.

The value of these collections of resources has been recognized in the research community, and they have been used to train systems to perform automatic fact-checking \cite{Popat:2017:TLE:3041021.3055133,wang:2017:Short,EMNLP2019:fauxtography} or to detect check-worthy claims in political debates \cite{Hassan:15,gencheva-EtAl:2017:RANLP,Patwari:17,RANLP2019:checkworthiness:multitask}.
There have also been datasets that combine claims from multiple fact-checking organizations \cite{augenstein-etal-2019-multifc}, again with the aim of performing automatic fact-checking.

\begin{figure}[tbh]
\centering
\includegraphics[trim={3cm 0 0 0},clip,width=0.95\columnwidth]{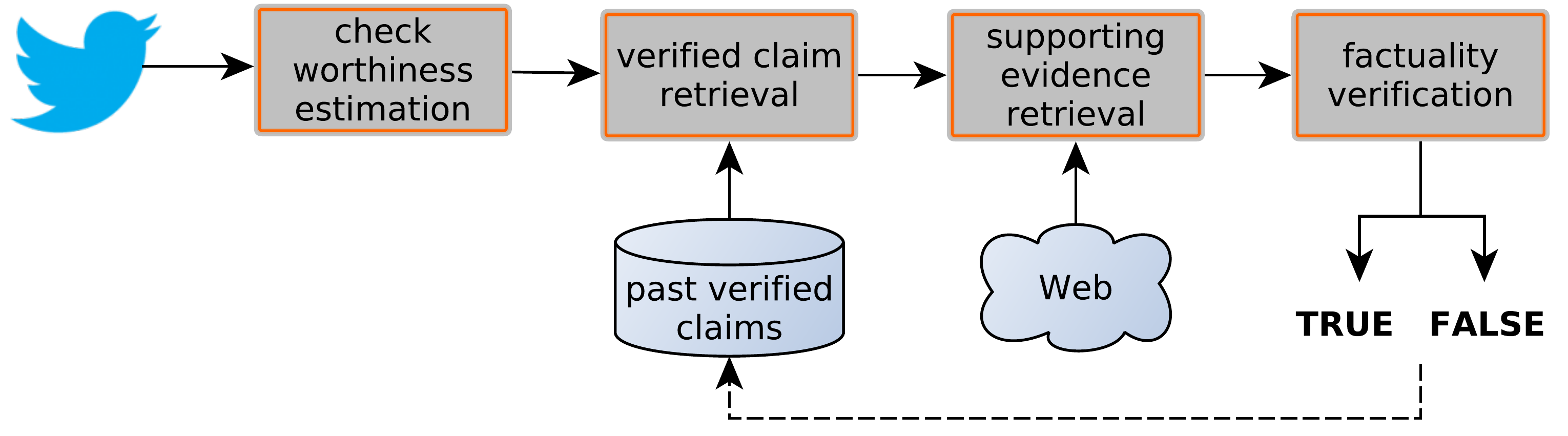}
\caption{A general information verification pipeline.}
\label{fig:pipeline}
\end{figure}

It has been argued that checking against a database of previously fact-checked claims should be an integral step of an end-to-end automated fact-checking pipeline \cite{Hassan:2017:CFE:3137765.3137815}. This is illustrated in Figure~\ref{fig:pipeline}, which shows the general steps of such a pipeline~\cite{CheckThat:ECIR2019}: (\emph{i})~assess the check-worthiness of the claim (which could come from social media, from a political debate, etc.), (\emph{ii})~check whether a similar claim has been previously fact-checked (the task we focus on here), (\emph{iii})~retrieve evidence (from the Web, from social media, from Wikipedia, from a knowledge base, etc.), and (\emph{iv})~assess the factuality of the claim.

From a fact-checkers' point of view, the abundance of previously fact-checked claims increases the likelihood that the next claim that needs to be checked would have been fact-checked already by some trusted organization. Indeed, viral claims often come back after a while in social media, and politicians are known to repeat the same claims over and over again.\footnote{President Trump has repeated one claim over 80 times: \url{http://tinyurl.com/yblcb5q5}.} Thus, before spending hours fact-checking a claim manually, it is worth first making sure that nobody has done it already.

On another point, manual fact-checking often comes too late. A study has shown that ``fake news'' spreads six times faster than real news~\cite{Vosoughi1146}. Another study has indicated that over 50\% of the spread of some viral claims happens within the first ten minutes of their posting on social media~\cite{zaman2014}.
At the same time, detecting that a new viral claim has already been fact-checked can be done automatically and very quickly, thus allowing for a timely action that can limit the spread and the potential malicious impact.

From a journalistic perspective, the ability to check quickly whether a claim has been previously fact-checked could be revolutionizing as it would allow putting politicians on the spot in real time, e.g.,~during a live interview. In such a scenario, automatic fact-checking would be of limited utility as, given the current state of technology, it does not offer enough credibility in the eyes of a journalist.

\noindent Interestingly, despite the importance of the task of detecting whether a claim has been fact-checked in the past, it has been largely ignored by the research community. Here, we aim to bridge this gap.
Our contributions can be summarized as follows:

\begin{itemize}
    \item We formulate the task and we discuss how it relates to, but differs from, previous work.
    \item We create a specialized dataset, which we release to the research community.\footnote{Data and code are available at the following URL:\\ \url{https://github.com/sshaar/That-is-a-Known-Lie}\label{footnote:published-dataset-link}} Unlike previous work in fact-checking, which used normalized claims from fact-checking datasets, we work with naturally occurring claims, e.g.,~in debates or in social media.
    \item We propose a learning-to-rank model that achieves sizable improvements over state-of-the-art retrieval and textual similarity models.
\end{itemize}

The remainder of this paper is organized as follows: Section~\ref{sec:related} discusses related work,  Section~\ref{sec:task} introduces the task, Section~\ref{sec:dataset} presents the dataset,
Section~\ref{sec:evaluationmeasures} discusses the evaluation measures,
Section~\ref{sec:models} presents the models we experiment with,
Section~\ref{sec:experiments} described our experiments, and Section~\ref{sec:conclusion} concludes and discusses future work. 

\section{Related Work}
\label{sec:related}

To the best of our knowledge, the task of detecting whether a claim has been previously fact-checked was not addressed before. \citet{Hassan:2017:CFE:3137765.3137815} mentioned it as an integral step of their end-to-end automated fact-checking pipeline, but there was very little detail provided about this component and it was not evaluated.

In an industrial setting, Google has developed \emph{Fact Check Explorer},\footnote{\url{http://toolbox.google.com/factcheck/explorer}} which is an exploration tool that allows users to search a number of fact-checking websites (those that use ClaimReview from \texttt{schema.org}\footnote{\url{http://schema.org/ClaimReview}}) for the mentions of a topic, a person, etc. However, the tool cannot handle a complex claim, as it runs Google search, which is not optimized for semantic matching of long claims. While this might change in the future, as there have been reports that Google has started using BERT in its search, at the time of writing, the tool could not handle a long claim as an input. 

\noindent A very similar work is the \emph{ClaimsKG} dataset and system~\cite{ClaimsKG}, which includes 28K claims from multiple sources, organized into a knowledge graph (KG). The system can perform data exploration, e.g.,~it can find all claims that contain a certain named entity or keyphrase. In contrast, we are interested in detecting whether a claim was previously fact-checked.

Other work has focused on creating datasets of textual fact-checked claims, without building KGs. Some of the larger ones include the \emph{Liar, Liar} dataset of 12.8K claims from PolitiFact \cite{wang:2017:Short}, and the \emph{MultiFC} dataset of 38K claims from 26 fact-checking organizations \cite{augenstein-etal-2019-multifc}, 
the 10K claims \emph{Truth of Various Shades} \cite{rashkin-etal-2017-truth} dataset, 
among several other datasets, which were used for automatic fact-checking of individual claims, not for checking whether an input claim was fact-checked previously. Note that while the above work used manually normalized claims as input, we work with naturally occurring claims as they were made in political debates and speeches or in social media.

There has also been a lot of research on automatic fact-checking of claims and rumors, going in several different directions. One research direction focuses on the social aspects of the claim and how users in social media react to it \cite{Canini:2011,Castillo:2011:ICT:1963405.1963500,ma2016detecting,gorrell-etal-2019-semeval,Ma:2019:DRT:3308558.3313741}. 
Another direction mines the Web for information that proves or disproves the claim \cite{mukherjee2015leveraging,RANLP2017:factchecking:external,Popat:2017:TLE:3041021.3055133,baly-EtAl:2018:N18-2,AAAI2018:factchecking,nadeem-etal-2019-fakta}. 
In either case, it is important to model the reliability of the source as well as the stance of the claim with respect to other claims; in fact, it has been proposed that a claim can be fact-checked based on its source alone~\cite{D18-1389} or based on its stance alone~\cite{dungs-EtAl:2018:C18-1}.
A third direction performs fact-checking against Wikipedia~\cite{thorne-etal-2018-fever,nie2019combining}, or against a general collection of documents~\cite{Miranda:2019:AFC:3308558.3314135}.
A fourth direction uses a knowledge base or a knowledge graph~\cite{10.1371/journal.pone.0128193,DBLP:conf/icdm/ShiralkarFMC17,Gad-Elrab:2019:EFE:3289600.3290996,Gad-Elrab:2019:TTF:3308558.3314126,Huynh:2019:BFC:3357384.3358036}.
Yet another direction performs fact-checking based on tables~\cite{2019TabFactA}.
There is also recent work on using language models as knowledge bases~\cite{petroni-etal-2019-language}.
Ours is yet another research direction.

\noindent While our main contribution here is the new task and the new dataset, we should also mentioned some work on retrieving documents. In our experiments, we perform retrieval using BM25~\cite{Robertson:2009:PRF:1704809.1704810} and re-ranking using BERT-based similarity, which is a common strategy in recent state-of-the-art retrieval models~\cite{akkalyoncu-yilmaz-etal-2019-applying,DBLP:journals/corr/abs-1901-04085,akkalyoncu-yilmaz-etal-2019-cross}. 

Our approach is most similar to that of \cite{akkalyoncu-yilmaz-etal-2019-applying}, but we differ, as we perform matching, both with BM25 and with BERT, against the normalized claim, against the title, and against the full text of the articles in the fact-checking dataset; we also use both scores and reciprocal ranks when combining different scores and rankings. Moreover, we use sentence-BERT instead of BERT. Previous work has argued that BERT by itself does not yield good sentence representation. Thus, approaches such as sentence-BERT \cite{reimers-gurevych-2019-sentence} have been proposed, which are specifically trained to produce good sentence-level representations. This is achieved using Siamese BERT networks that are fine-tuned on NLI and STS-B data. Indeed, in our experiments, we found sentence-BERT to perform much better than BERT. The Universal Sentence Encoder~\cite{cer-etal-2018-universal} is another alternative, but sentence-BERT worked better in our experiments.

Finally, our task is related to semantic relatedness tasks, e.g.,~from the GLUE benchmark \citep{wang-etal-2018-glue}, such as natural language inference, or NLI task \cite{williams-etal-2018-broad}, recognizing textual entailment, or RTE \cite{Bentivogli09thefifth}, paraphrase detection~\cite{dolan-brockett-2005-automatically}, and semantic textual similarity, or STS-B \cite{cer-etal-2017-semeval}. However, it also differs from them, as we will see in the following section.

\begin{table*}[tbh]
  \small
  \centering
  \begin{tabular}{cp{0.45\linewidth}p{0.45\linewidth}}
    \toprule
    \bf No. & \multicolumn{1}{c}{\bf \Inputclaim~claim} & \multicolumn{1}{c}{\bf Manually annotated corresponding claim in PolitiFact} \\
    \midrule
    1 & \emph{Richard Nixon released tax returns when he was under audit.} & Richard Nixon released tax returns when he was under audit. \\
    \hline
    2 & \emph{Hillary wants to give amnesty.} & Says Hillary Clinton ``wants to have open borders.'' \\
    \hline
    3 & \emph{People with tremendous medical difficulty and medical problems are pouring in, and in many, in many cases, it's contagious.} & Says ``2,267 caravan invaders have tuberculosis, HIV, chickenpox and other health issues'' \\
    \hline
    4 & \emph{He actually advocated for the actions we took in Libya and urged that Gadhafi be taken out, after actually doing some business with him one time.} & Says Donald Trump is ``on record extensively supporting the intervention in Libya.'' \\
    \hline
    5 & \emph{He actually advocated for the actions we took in Libya and urged that Gadhafi be taken out, after actually doing some business with him one time.} & When Moammar Gadhafi was set to visit the United Nations, and no one would let him stay in New York, Trump allowed Gadhafi to set up an elaborate tent at his Westchester County (New York) estate. \\
    \bottomrule
  \end{tabular}
  \caption{\textbf{PolitiFact:} \emph{\Inputclaim}--\emph{\verified}~claim pairs. The \inputclaim~claims are sentences from the 2016 US Presidential debates, and the \verified~claims are their corresponding fact-checked counter-parts in PolitiFact.}
  \label{table:claim-fact-dataset}
\end{table*}

\section{Task Definition}
\label{sec:task}

We define the task as follows: 
\textit{Given a check-worthy input claim and a set of verified claims, rank those verified claims, so that the claims that can help verify the input claim, or a sub-claim in it, are ranked above any claim that is not helpful to verify the input claim.}

Table~\ref{table:claim-fact-dataset} shows some examples of \emph{\inputclaim--\verified} claim pairs, where the \inputclaim~claims are sentences from the 2016 US Presidential debates, and the \verified~claims are the corresponding fact-checked counter-parts in PolitiFact.

We can see on \mbox{line 1} of Table~\ref{table:claim-fact-dataset} a trivial case, where the \verified~claim is identical to the \inputclaim~claim; however, such cases are not very frequent, as the experiments with the BM25 baseline in Section~\ref{sec:experiments} below will show. 

Lines 2 and 3 show harder cases, where the \inputclaim~claim and its manually annotated counter-part are quite different in their lexical choice, and yet the latter can serve to verify the former. 

Lines 4 and 5, show a complex \inputclaim~claim, which contains two sub-claims, each of which is verified by two corresponding claims in PolitiFact.

From the above examples, it is clear that ours is not a paraphrasing task, as illustrated by examples 2--5. It is also not a natural language inference (NLI) or a recognizing textual entailment (RTE) task, as a claim can have sub-claims, which complicates entailment reasoning (as illustrated by examples 4--5). Finally, the task goes beyond simple textual similarity, and thus it is not just an instance of semantic textual similarity (STS-B).

Note that we do not try to define formally what makes a \verified~claim a good match for an \inputclaim~claim. Instead, we trust the manual annotations for this by fact-checking experts, which they perform when they comment on the claims made in political debates and speeches. In many cases, the fact-checkers have explicitly indicated which previously fact-checked claim corresponds to a given original claim in a debate/speech. A similar approach was adopted for a related task, e.g.,~it was used to obtain annotated training and testing data for the Check-Worthiness task of the CLEF CheckThat! Lab \cite{clef2018checkthat:task1,clef-checkthat-T1:2019,ECIR2020:CLEF}.

\section{Datasets}
\label{sec:dataset}

We created two datasets by collecting, for each of them, a set of \emph{\verified~claims} and matching \emph{\inputclaim--\verified}~claims pairs (below, we will also refer to these pairs as \textit{\Inputclaim-\verclaim} pairs): the first dataset, PolitiFact, is about political debates and speeches and it is described in Section~\ref{sec:politifacts}; the second dataset, Snopes, includes tweets, and it is described in Section~\ref{sec:snopes}.

\subsection{PolitiFact Dataset}\label{sec:politifacts}

PolitiFact is a fact-checking website that focuses on claims made by politicians, elected officials, and influential people in general. 
PolitiFact fact-checks claims by assigning a truth value to them and publishing an article that gives background information and explains the assigned label. This is similar to how other fact-checking websites operate.
 
We retrieved a total of 16,636 \verified~claims from PolitiFact, populating for each of them the following fields:

\begin{itemize}
    \item \textit{\verclaim}: the text of the claim, which is a normalized version of the original claim, as the human fact-checkers typically reformulate it, e.g.,~to make it clearer, context-independent, and self-contained;
    \item \textit{TruthValue}: the label assigned to the claim;\footnote{We do not use the claim veracity labels in our experiments, but we collect them for possible future use.}
    \item \textit{Title}: the title of the article on PolitiFact that discusses the claim;
    \item \textit{Body}: the body of the article.
\end{itemize}

\begin{table*}
  \small
  \centering
  \begin{tabular}{cp{0.45\linewidth}p{0.45\linewidth}}
    \toprule
    \bf No. & \multicolumn{1}{c}{\bf Tweet} & \multicolumn{1}{c}{\bf Manually annotated corresponding claim in Snopes} \\
    \midrule
    1 & \emph{Welp\ldots it’s official\ldots Kim Kardashian finally decided to divorce Kanye West… https://t.co/C2p25mxWJO --- Ashlee Marie Preston (@AshleeMPreston) October 12, 2018} & Kanye West and Kim Kardashian announced that they were divorcing in October 2018. \\
    \hline
    2 & \emph{Kim Kardashian and Kanye West are splitting up https://t.co/epwKG7aSBg pic.twitter.com/u7qqojWVlR --- ELLE Magazine (US) (@ELLEmagazine) October 18, 2018} & Kanye West and Kim Kardashian announced that they were divorcing in October 2018. \\
    \hline
    3 & \emph{Everyone should be able to access high-quality, affordable, gender-affirming health care. But the Trump administration is trying to roll back important protections for trans Americans. Help fight back by leaving a comment for HHS in protest: https://t.co/pKDcOqbsc7 --- Elizabeth Warren (@ewarren) August 13, 2019} & U.S. Sen. Elizabeth Warren said or argued to the effect that taxpayers must fund sex reassignment surgery. \\
    \bottomrule
  \end{tabular}
  \caption{\textbf{Snopes:} \emph{\Inputclaim}--\emph{\verclaim}~claim pairs. The \inputclaim~claims are tweets and the \verified~claims are their corresponding fact-checked counter-parts in Snopes.}
  \label{table:snopes-dataset}
\end{table*}

Often, after a major political event, such as a political speech or a debate, PolitiFact publishes reports\footnote{Note that these reports discuss multiple claims, unlike the typical PolitiFact article about a specific claim.} that discuss the factuality of some of the claims made during that event. 
Importantly for us, in these reports, some of the claims are linked to previously verified claims in PolitiFact. Such pairs of an original claim and a previously verified claim form our \textit{Claim--\verclaim}~pairs. 

We collected such overview reports for 78 public events in the period 2012--2019, from which we collected a total of 768 \emph{\Inputclaim--\verclaim}~pairs. 
Given an \emph{\Inputclaim}~claim, we refer to the corresponding \emph{\verified}~claim in the pair as its \emph{matching \verclaim~claim}. In general, there is a 1:1 correspondence, but in some cases an \emph{\Inputclaim}~claim is mapped to multiple \emph{\verclaim}~claims in the database, and in other cases, multiple \emph{\Inputclaim}~claims are matched to the same \emph{\verclaim}~claim.

Thus, the task in Section~\ref{sec:task} reads as follows when instantiated to the PolitiFact dataset: given an \emph{\Inputclaim}~claim, rank all 16,636 \emph{\verclaim}~claims, so that its \emph{matching \verclaim}~claims are ranked at the top.

\subsection{Snopes Dataset}
\label{sec:snopes}

Snopes is a website specialized in fact-checking myths, rumors, and urban legends. We used information from it to create a second dataset, this time focusing on tweets. We started with a typical article about a claim, and we looked inside the article for links to tweets that are possibly making that claim. Note that some tweets mentioned in the article are not making the corresponding \verified~claim, and some are not making any claims; we manually checked and filtered out such tweets.

We collected 1,000 suitable tweets as \emph{\Inputclaim}~claims, and we paired them with the corresponding claim that the page is about as the \emph{\verclaim}~claim.
We further extracted from the article its \textit{Title}, and the \textit{TruthValue} of the \Inputclaim~claim (a rating of the claims assigned from Snopes\footnote{\url{http://www.snopes.com/fact-check-ratings/}}). 

Examples of \emph{\inputclaim--\verclaim}~pairs are shown in Table~\ref{table:snopes-dataset}. 
Comparing them to the ones from Table~\ref{table:claim-fact-dataset}, we can observe that the Snopes tweets are generally more self-contained and context-independent.

Finally, we created a set of \emph{\verclaim}~claims to match against using the Snopes claims in the \emph{ClaimsKG} dataset~\cite{ClaimsKG}.
Ultimately, our Snopes dataset consists of 1,000 \emph{\inputclaim--\verclaim}~pairs and 10,396 verified claims. 

Statistics about the datasets are shown in Table~\ref{table:datasets:statistics}; the datasets are available online.\textsuperscript{\ref{footnote:published-dataset-link}}

\subsection{Analysis}

In section~\ref{sec:task}, we discussed that matching some of the \inputclaim~claims with the corresponding \verified~claims can be a non-trivial task, and we gave examples of easy and hard cases. 
To capture this distinction, we classify \Inputclaim--\verclaim~ pairs into two types.
\textit{Type-1} pairs are such for which the \Inputclaim~claim can be matched to the \verclaim~using simple approximate string matching techniques., e.g., as in line 1 of Table~\ref{table:claim-fact-dataset} and lines 1-2 of Table~\ref{table:snopes-dataset}. 
Conversely, \textit{Type-2} pairs are such for which the \Inputclaim~claim cannot be easily mapped to the \verclaim, e.g., as in lines 2-5 of Table~\ref{table:claim-fact-dataset} and line 3 of Table~\ref{table:snopes-dataset}. 
We manually annotated a sample of 100 pairs from PolitiFact \inputclaim--\verclaim pairs and we found 48\% of them to be of \textit{Type-2}.

We further analyzed the complexity of matching an \emph{\Inputclaim}~claim to the \emph{\verclaim} from the same \emph{\Inputclaim--\verclaim} pair using word-level TF.IDF-weighted cosine similarity. Table~\ref{table:cosine-similarity} shows the number of pairs for which this similarity is above a threshold. 
We can see that, for PolitiFact, only 27\% of the pairs have a similarity score that is above 0.25, while for Snopes, this percentage is at 50\%, which suggests Snopes should be easier than PolitiFact.

\begin{table}[t]
    \centering
    \begin{tabular}{l@{ }@{ }@{ }rr}
        \toprule
        & \bf PolitiFact & \bf Snopes\\        
        \midrule
        \emph{\Inputclaim}--\emph{\verclaim}~pairs & 768 & 1,000\\
        -- training & 614 & 800\\
        -- testing & 154 & 200\\        
        Total \# of verified claims & 16,636 & 10,396\\
        \bottomrule
    \end{tabular}
    \caption{\textbf{Statistics about the datasets:} shown are the number of \emph{\Inputclaim}--\emph{\verclaim}~pairs and the total number of \emph{\verclaim}~claims to match an \emph{\Inputclaim}~claim against. Note that each  \emph{\verclaim} comes with an associated fact-checking analysis document in PolitiFact/Snopes.}
  \label{table:datasets:statistics}
\end{table}

\begin{table}[t]
    \centering
    \begin{tabular}{crrrr}
            \toprule 
            & \multicolumn{2}{c}{\bf PolitiFact} & \multicolumn{2}{c}{\bf Snopes}\\
            \midrule 
            \bf Threshold & \bf \#  & \bf \% & \bf \# & \bf \% \\
            \midrule
            0.75 & 55 & $8\%$ & 11 & 1\%\\
            0.50 & 128 & $17\%$ & 75 & 8\%\\
            0.25 & 201 & $27\%$ & 504 & 50\%\\            
            \midrule
            0.00 & 768 & 100\% & 1,000 & 100\%\\
            \bottomrule
        \end{tabular}
    \caption{\textbf{Analysis of the task complexity:} number of \emph{\Inputclaim}--\emph{\verclaim}~pairs in PolitiFact and Snopes with TF.IDF-weighted cosine similarity above a threshold.}
  \label{table:cosine-similarity}
\end{table}

\begin{table*}[tbh]
    \small
    \centering
    \begin{tabular}{@{}l@{}c*{12}c@{}}
        \toprule
        \bf Experiment & \multicolumn{1}{c}{\bf MRR} & \multicolumn{6}{c}{\bf MAP@$k$} & \multicolumn{6}{c}{\bf HasPositives@$k$} \\
        \cmidrule(lr){3-8}\cmidrule(lr){9-14}
        {} & {} & 1 & 3 & 5 & 10 & 20 & all & 1 & 3 & 5 & 10 & 20 & 50 \\
        \midrule 
        \multicolumn{14}{c}{\bf IR (BM25; full database)}\\
        \midrule 
        IR:Title	&	.288	&	.216	&	.259	&	.261	&	.268	&	.272	&	.276	&	.220	&	.330	&	.346	&	.401	&	.464	&	.535 \\
        IR:\verclaim	&	.435	&	.366	&	.404	&	.413	&	.415	&	.419	&	.422	&	.378	&	.472	&	.511	&	.527	&	.574	&	.629 \\
        IR:Body	&	\bf.565	&	\bf.484	&	\bf.538	&	\bf.546	&	\bf.552	&	\bf.556	&	\bf.560	&	\bf.488	&	\bf.614	&	\bf.653	&	\bf.700	&	\bf.740	&	\bf.811 \\
        IR:Title+\verclaim+Body	&.526	&	.425	&	.504	&	.507	&	.513	&	.516	&	.519	&	.433	&	\bf.614	&	.630	&	.661	&	.717	&	.772 \\
        \midrule 
        \multicolumn{14}{c}{\bf Semantic Matching - BERT \& co. (matching against \verclaim~only; full db)}\\
        \midrule
        BERT:base,uncased	&	.268	&	.204	&	.242	&	.251	&	.259	&	.260	&	.264	&	.204	&	.299	&	.338	&	.393	&	.409	&	.496 \\
		RoBERTa:base	&	.209	&	.173	&	.194	&	.198	&	.203	&	.205	&	.207	&	.173	&	.220	&	.236	&	.283	&	.315	&	.346 \\
		sentence-BERT:base	&	.377	&	.311	&	.352	&	.354	&	.361	&	.366	&	.370	&	.315	&	\bf.417	&	\bf.425	&	\bf.480	&	\bf.551	&	\bf.614 \\
		sentence-BERT:large	&	\bf.395	&	\bf.354	&	\bf.367	&	\bf.372	&	\bf.381	&	\bf.382	&	\bf.386	&	\bf.362	&	.393	&	.417	&	\bf.480	&	.496	&	.582 \\
        \midrule 
        \multicolumn{14}{c}{\bf BERT on Full Articles (sent.BERT:large matching against \verclaim~+ Title + top-$n$ article body sent.; full db)}\\
        \midrule        
        sentence-BERT: n = 3	&	.515	&	\bf.441	&	.478	&	.493	&	.498	&	.501	&	.505	&	\bf.457	&	.528	&	\bf.598	&	\bf.638	&	\bf.693    &    \bf.756 \\
        sentence-BERT: n = 4	&	\bf.517	&	\bf.441	&	\bf.487	&	\bf.497	&	\bf.500	&	\bf.505	&	\bf.508	&	\bf.457	&	\bf.551	&	\bf.598	&	.622	&	.685    &   \bf.756 \\
        sentence-BERT: n = 5	&	.515	&	.433	&	.484	&	.491	&	.498	&	.502	&	.505	&	.449	&	\bf.551	&	.583	&	\bf.638	&	.685    &    .748 \\
        sentence-BERT: n = 6	&	.509	&	.429	&	.480	&	.485	&	.491	&	.497	&	.500	&	.441	&	.543	&	.567	&	.614	&	\bf.693    &    .740 \\
        \midrule 
        \multicolumn{14}{c}{\bf Reranking (IR \& sent.BERT:large matching against \verclaim~+ Title + top-4 article body sent.; top-$N$ from IR)}\\
        \midrule        
        Rerank-IR-top-10	&	.586	&	.528	&	.572	&	.578	&	.583	&	---	&	---	&	.512	&	.638	&	.693	&	.701	&	---	&	--- \\
        Rerank-IR-top-20	&	.586	&	.521	&	.572	&	.577	&	.580	&	.583	&	---	&	.512	&	.646	&	.685	&	.709	&	.740	&	--- \\
        Rerank-IR-top-50	&	.600	&	.519	&	.568	&	.584	&	.590	&	.590	&	.594	&	.520	&	.638	&	\bf.717	&	\bf.772	&	.780	&	.811 \\
        Rerank-IR-top-100	&	\bf.608	&	\bf.531	&	\bf.580	&	\bf.588	&	\bf.597	&	\bf.599	&	\bf.602	&	\bf.535	&	\bf.654	&	.685	&	.740	&	\bf.787	&	\bf.819 \\
        Rerank-IR-top-200	&	.605	&	.529	&	.575	&	.585	&	.594	&	.598	&	.599	&	\bf.535	&	.646	&	.685	&	.756	&	.803	&	.811 \\
        \bottomrule
    \end{tabular}
    \caption{\textbf{PolitiFact:} evaluation results on the test set.}
    \label{table:results-politifact}
\end{table*}

\begin{table*}[tbh]
    \small
    \centering
    \begin{tabular}{@{}l@{}c*{12}c@{}}
        \toprule
        \bf Experiment & \multicolumn{1}{c}{\bf MRR} & \multicolumn{6}{c}{\bf MAP@$k$} & \multicolumn{6}{c}{\bf HasPositives@$k$} \\
        \cmidrule(lr){3-8}\cmidrule(lr){9-14}
        {} & {} & 1 & 3 & 5 & 10 & 20 & all & 1 & 3 & 5 & 10 & 20 & 50 \\
        \midrule 
        \multicolumn{14}{c}{\bf IR (BM25; full database)}\\
        \midrule 
         IR:Title       &  .619  &  .538  &  .573  &  .583  &  .587  &  .590  &  .592  &  .538  &  .673  &  .724  &  .759  &  .804  &  .844  \\
        IR:\verclaim  &  .655  & \bf .555 & .589 & .598 & .600 & .602 & .605 & \bf .558 & .729 & .769 & .784 & .809 & .864 \\
        IR:\verclaim+Title  &  \bf .664  & \bf .555 & \bf .592 & \bf .600 & \bf .605 & \bf .608 & \bf .609 & \bf .558 & \bf .739 & \bf .774 & \bf .814 & \bf.849 & \bf .879  \\
        \midrule 
        \multicolumn{14}{c}{\bf Semantic Matching - BERT \& co. (matching against \verclaim~only; full database)}\\
        \midrule
        sent.BERT-base:Title  & .474  & .397 & .417 & .425 & .430 & .434 & .437 & .397 & .528 & .563 & .598 & .658 & .729  \\
        sent.BERT-base:\verclaim  & .515  & .402 & .489 & .504 & .510 & .512 & .515 & .402 & .593 & \bf .653 & \bf .698 & \bf .739 & .784  \\
        sent.BERT-large:\verclaim  & \bf .543 & \bf .457 & \bf .518 & \bf .527 & \bf .533 & \bf .535 & \bf .538 & \bf .457 & \bf .603 & .648 & .693 & .724 & \bf .794  \\
        \midrule 
        \multicolumn{14}{c}{\bf Reranking (IR \& sentence-BERT:large matching against \verclaim~+ Title; top-$N$ from IR)}\\
        \midrule
        Rerank-IR-top-10  & .764 & .687 & .762 & .764 & .764 & --- & --- & .673 & .859 & .869 & .869 & --- & --- \\
        Rerank-IR-top-20  & .781 & .686 & .773 & .780 & .781 & .781 & --- & .678 & .869 & \bf .905 & \bf .920 & .920 & --- \\
        Rerank-IR-top-50  & \bf .788 & \bf .691 & \bf .780 & \bf .782 & \bf .784 & \bf .784 & \bf .787 & \bf .693 & \bf .874 & .894 & .915 & .925 & .930 \\
        Rerank-IR-top-100  & .775 & .669 & .758 & .760 & .760 & .760 & .774 & .673 & .859 & .889 & .910 & .925 & .930 \\        
        Rerank-IR-top-200 & .778 & .672 & .762 & .764 & .764 & .764 & .777 & .678 & .864 & .884 & .910 & \bf .930 & \bf .950 \\        
        \bottomrule 
    \end{tabular}
    \caption{\textbf{Snopes:} evaluation results on the test set.}
    \label{table:snopes-results}
\end{table*}

\section{Evaluation Measures} \label{sec:evaluationmeasures}

We treat the task as a ranking problem. Thus, we use ranking evaluation measures, namely mean reciprocal rank (MRR), Mean Average Precision (MAP), and MAP truncated to rank $k$ (MAP@$k$). We also report HasPositive@$k$, i.e.,~whether there is a true positive among the top-$k$ results.

Measures such as MAP@$k$ and HasPositive@$k$ for $k \in \{1,3,5\}$ would be relevant in a scenario, where a journalist needs to verify claims in real time, in which case the system would return a short list of 3-5 claims that the journalist can quickly skim and make sure they are indeed a true match.

We further report MAP@$k$ and HasPositive@$k$ for $k \in \{10,20\}$ as well as MAP (untruncated), which would be more suitable in a non-real-time scenario, where recall would be more important.

\section{Models} \label{sec:models}

Here, we describe the models we experiment with. 

\subsection{BM25} \label{sec:bm25}

A simple baseline is to use BM25~\cite{Robertson:2009:PRF:1704809.1704810}, which is classical approach in information retrieval. BM25 assigns a score to each query-document pair based on exact matching between the words in the query and the words in a target document, and it uses this score for ranking. We experiment with BM25 using the input claim as a query against different representations of the verified claims:

\begin{itemize}
    \item \textbf{IR (Title):} the article titles;
    \item \textbf{IR (\verclaim):} the verified claims;
    \item \textbf{IR (Body):} the article bodies;
    \item Combinations of the above.
\end{itemize}

\subsection{BERT-based Models} \label{sec:bert}

The BM25 algorithm focuses on exact matches, but as lines 2--5 in Table~\ref{table:claim-fact-dataset} and  line 3 in Table~\ref{table:snopes-dataset} show, the input claim can use quite different words. Thus, we further try semantic matching using BERT.

Initially, we tried to fine-tune BERT~\cite{devlin-etal-2019-bert}, but this did not work well, probably because we did not have enough data to perform the fine-tuning.
Thus, eventually we opted to use BERT (and variations thereof) as a sentence encoder, and to perform max-pooling on the penultimate layer to obtain a representation for an input piece of text.
Then, we calculate the cosine similarity between the representation of the input claim and of the \verified~claims in the dataset, and we use this similarity for ranking. 

\begin{itemize}
    \item \textbf{BERT:base,uncased}: the base, uncased model of BERT;
    \item \textbf{RoBERTa:base}: the base, cased model of RoBERTa~\cite{roberta};
    \item \textbf{sentence-BERT:base}: BERT, specifically trained to produce good sentence representations ~\cite{reimers-gurevych-2019-sentence}; this is unlike BERT and RoBERTa, for which we found the cosine similarity between totally unrelated claims often to be quite high;
    \item \textbf{sentence-BERT:large}: the large version of sentence-BERT.
    \item \textbf{BERT on full articles:} We further extend the above models to match against the body of the document, borrowing and further developing an idea from  \cite{yang2019simple}. We use sentence-BERT to encode each sentence in the \textit{Body}, and then we compute the cosine similarity between the input claim and each of those sentences. Next, we collect scores for each claim-document pair, as opposed to having only a single score representing the similarity between the input and a \verified~claim. These scores include the cosine similarity for (\emph{i})~claim vs. \textit{\verclaim}, (\emph{ii})~claim vs. \textit{Title}, and (\emph{iii})~top-$n$ scores of the claim vs. \textit{Body} sentences. Finally, we train a binary classifier that takes all these scores and predicts whether the claim-document pair is a good match.
\end{itemize}

\subsection{Reranking}\label{sec:reranking}

Since BM25 and BERT capture different types of information, they can be combined to create a set of features based on the rankings returned by BM25 and the similarity scores computed on the embedding of the claim pairs. 
Following~\cite{nogueira2019multistage}, we use a reranking algorithm, namely rankSVM with an RBF kernel, which learns to rank using a pairwise loss.

\section{Experiments}
\label{sec:experiments}

Below we describe our experiments on the PolitiFact and the Snopes datasets. We start with IR-based models, followed by BERT-based semantic similarity on claims and articles, and finally we experiment with pairwise learning-to-rank models.

\subsection{Politifact Experiments}\label{sec:experiments-politifact}

For the PolitFact dataset, we perform experiments with all models from Section~\ref{sec:models}, and we report the results in Table~\ref{table:results-politifact}.

\subsubsection{Experiment 1: BM25-based Baselines}\label{sec:information-retrieval.setup}

We ran experiments matching the \emph{\Inputclaim} against \textit{Title}, \textit{\verclaim}, \textit{Body} and \textit{Title+\verclaim+Body}. We can see in Table~\ref{table:results-politifact} that using the \textit{Title} yields the lowest results by a large margin. This is because the \textit{Title} is only a summary, while \textit{\verclaim} and \textit{Body} contain more details and context.
We can further see that the best representation, on all measures, is to use the \textit{Body}, which performs better than using \textit{\verclaim} by 0.12-0.14 in terms of MAP@$k$ and MAP, and by 0.09 on MRR. This is probably because the article body is longer, which increases the probability of having more words matching the input claim. Finally, matching against all three targets is slightly worse than using \textit{Body} only.

\subsubsection{Experiment 2: Semantic Matching}\label{sec:sementic-matching}

Next, we experimented with cosine similarity between the \emph{\Inputclaim}~claim and \textit{\verclaim}, as the BM25 experiments above have shown that using \emph{\verclaim}~is better than using \textit{Title}.

We can see in Table~\ref{table:results-politifact} that  BERT:uncased is better than RoBERTa (which is case sensitive) on all measures, which suggests that casing might not matter.
We further see that the best semantic model is sentence-BERT: both the base and the large variants of sentence-BERT beat BERT and RoBERTa by at least 13\% absolute across all measures (and in some cases, by a much larger margin).

\subsubsection{Experiment 3: BERT on Full Articles}\label{sec:bertarticle}

Next, we performed full article experiments, where we used the large model of sentence-BERT, as it outperformed the rest of the BERT models shown in Table~\ref{table:results-politifact}.
We extracted similarity scores for each claim-document pair using sentence-BERT:large. We then trained a simple neural network (20-relu-10-relu) for classification. We trained the model for 15 epochs with a batch size of 2,048 using the Adam optimizer with a learning rate of 1e-3. We further used class weighting because the data was severely imbalanced: there were 614 positive exampled out of 10M claim-document pairs, as we paired each of the 614 input claims with each of the 16,636 verified claims in the database.

We ran the experiment for various numbers of top-$n$ cosine scores obtained from the \textit{Body}, as we wanted to investigate the relationship between the model performance and the information it uses.

In the \emph{BERT on Full Articles} section in Table~\ref{table:results-politifact}, we can see that using the scores for the top-4 best-matching sentences from the article body, together with scores for \emph{\verclaim} and for the article title, yielded the best performance. Moreover, the results got closer to those for BM25, even though overall they still lag a bit behind.

\subsubsection{Experiment 4: Reranking}\label{sec:ranksvmresults}

Finally, we trained a pairwise RankSVM model to re-rank the top-$N$ results retrieved using \emph{IR:Body}. For each claim-document pair in the top-$N$ list, we collected the scores for \emph{IR:Title}, \emph{IR:\verclaim}, \emph{IR:Body}, as well as from sentence-BERT:large for $n=4$ with their corresponding reciprocal ranks for the rankings they induce. As described in Section~\ref{sec:reranking}, using both methods yields better predictions as this combines exact matching and semantic similarities. 

We can see in Table~\ref{table:results-politifact} that the re-ranker yielded consistent and sizable improvement over the models from the previous experiments, by 0.04-0.05 points absolute across the different measures, which is remarkable as it is well-known from the literature that BM25 is a very strong baseline for IR tasks.
This is because our reranker is able to use both exact and semantic matching to target the different kinds of pairs that are found in the dataset. 
We also notice that the performance of the re-ranker improves as we increase the length of the list that is being re-ranked until a length of 100, and it starts degrading after that.


\subsection{Experiments on Snopes}
\label{sec:experiments-snopes}

On the Snopes dataset, we performed experiments analogous to those for the PolitiFact dataset, but with some differences, the most important being that this time we did not perform matching against the article body as the tweets that serve as input claims in our Snopes dataset were extracted from the article body. Note that this was not an issue for the PolitiFact dataset, as the input claim in a debate/speech required a lot of normalization and could not be found in the article body verbatim.
Table~\ref{table:snopes-results} reports the evaluation results.

\subsubsection{Experiment 1: BM25-based Baselines} 
\label{sec:information-retrieval-snopes}

We ran three experiments using BM25 to match the \emph{\Inputclaim} against \textit{Title}, \textit{\verclaim}, and \textit{Title+\verclaim}.
We can see in Table~\ref{table:snopes-results} that, just like for PolitiFact, using \emph{\verclaim}~performed better than using the article title, which is true for all evaluation measures; however, this time the margin was much smaller than it was for PolitiFact.
We further noticed a small improvement for all MAP@$k$ measures when matching against both the article \textit{Title} and the \emph{\verclaim}. Overall, BM25 is a very strong baseline for Snopes due to the high word overlap between the input claims and the verified claims (also, compared to PolitiFact, as we have seen in Table~\ref{table:cosine-similarity} above).

\subsubsection{Experiment 2: Semantic Matching}
\label{sec:sementic-matching-snopes}

Based on the lessons learned from PolitiFact, for semantic matching, we only experimented  with sentence-BERT.
We can see in Table~\ref{table:snopes-results} that this yielded results that were lower than for BM25 by a margin of at least 0.10 absolute for almost every reported measure; yet, this margin is smaller than for PolitiFact.
For these experiments, once again matching against the verified claim outperformed matching against the article title by a sizable margin.

\subsubsection{Experiment 3: BERT on Full Articles}
\label{sec:bertarticle-snopes}

As mentioned above, we did not perform matching of the input tweet against the article body, as this would easily give away the answer: the tweet can be found verbatim inside the target article.

For the purpose of comparison, we tried to filter out the text of the input tweet from the text of the article body before attempting the matching, but we still got unrealistically high results. Thus, ultimately we decided to abandon these experiments.

\subsubsection{Experiment 4: Reranking}\label{sec:ranksvm-results-snopes}

Finally, we trained a pairwise RankSVM model to re-rank the top-$N$ results from \emph{IR:VerClaim+Title}. 
For each claim-document pair in the top-$N$ list, we extracted the scores from \textit{IR:Title}, \textit{IR:VerClaim}, \textit{IR:VerClaim+Title}, \emph{sentence-BERT:large:Title}, and \emph{sentence-BERT:large:VerClaim}, as well as the corresponding reciprocal ranks for all target documents according to each of these scores.
This is the same as for PolitiFact, except that now we do not use scores for matching the input to a document body.
We can see in Table~\ref{table:snopes-results} that the best re-ranking model yielded sizable improvements over the best individual model by 0.09-0.18 points absolute on all evaluation measures.

Comparing the best re-ranking models for Snopes and PolitiFact, we can see that Snopes performed best when using a top-50 list, compared to top-100 for PolitiFact. 
We believe that this is due to the difference in performance of the retrieval models used to extract the top-$N$ pairs: for Snopes, \emph{IR:VerClaim+Title} has an MMR score of 0.664, while the best PolitiFact model, \emph{IR:Body}, has an MRR score of 0.565. Thus, for Snopes we rerank an $N$-best list extracted by a stronger IR model, and thus there is no need to go that deep in the list.

\section{Conclusions and Future Work}
\label{sec:conclusion}

We have argued for the need to address detecting previously fact-checked claims as a task of its own right, which could be an integral part of automatic fact-checking, or a tool to help human fact-checkers or journalists. We have created specialized datasets, which we have released, together with our code, to the research community in order to enable further research. 
Finally, we have presented learning-to-rank experiments, demonstrating sizable improvements over state-of-the-art retrieval and textual similarity approaches.

In future work, we plan to extend this work to more datasets and to more languages. We further want to go beyond textual claims, and to take claim-image and claim-video pairs as an input.

\section*{Acknowledgments}
This research is part of the Tanbih project,\footnote{\url{http://tanbih.qcri.org/}} which aims to limit the effect of `fake news,'' disinformation, propaganda, and media bias by making users aware of what they are reading.

\bibliography{acl2020}
\bibliographystyle{acl_natbib}

\end{document}